\documentclass{article}
\usepackage{spconf,amsmath,graphicx,amssymb,algorithm,algpseudocode,multirow,hyperref,float,comment}

\title{Towards Audio to Scene Image Synthesis \\ using Generative Adversarial Network}
\name{Chia-Hung Wan$^1$, Shun-Po Chuang$^2$, Hung-Yi Lee$^2$}
\address{$^1$Graduate Institute of Electrical Engineering, National Taiwan University\\$^2$Graduate Institute of Communication Engineering, National Taiwan University}

\begin{document}
%
\maketitle
\begin{abstract}
Humans can imagine a scene from a sound.
We want machines to do so by using conditional generative adversarial networks (GANs).
By applying the techniques including spectral norm, projection discriminator and auxiliary classifier, compared with naive conditional GAN, the model can generate images with better quality in terms of both subjective and objective evaluations.
Almost three-fourth of people agree that our model have the ability to generate images related  to sounds.
By inputting different volumes of the same sound, our model output different scales of changes based on the volumes, showing that our model truly knows the relationship between sounds and images to some extent.
\end{abstract}



%

\section{Introduction}
\label{sec:intro}

People now are trying to make machines work like humans.
Researchers are attempting to teach machines to comprehend natural languages, to understand the content in images, etc.
After understanding the content, we also want machines to describe what they see.
For example, in video caption generation~\cite{chuang2017seeing}, machine describes what contents are in the video after watching it.
In addition, we also want machines to have the ability to imagine.
In the task of text-to-image\cite{reed2016generative}, machine can turn text descriptions into images.
In this paper, we want machines to imagine the scenes by listening to sounds.
We hope that when hearing sounds, machine can draw the object that is making sounds and the scene that the sound is made.
For instance, after hearing the sparrows crow, machine can draw a picture of sparrows with probably trees or grass as background.

In recent years, there are lots of generative models using generative adversarial networks (GANs)~\cite{goodfellow2014generative} to generate images. 
Besides generating images randomly, there is also a large number of researches using conditional GANs~\cite{mirza2014conditional}, in which the generators take some conditions as input and generate corresponding images.
In the previous work, their conditions are the text description of images~\cite{reed2016generative} or the classes of the images to be generated\cite{miyato2018cgans}\cite{odena2016conditional}.
Based on conditional GANs, if we can provide enough sounds and their corresponding images, machines are supposed to learn how to generate images that include the objects making sounds. 
As far as we know, there is little image generative model that is conditioned on sound. 

The technology we use to learn an audio-to-image generator is based on GAN.
In this paper, we fuse several advanced techniques of conditional GANs including spectral normalization~\cite{miyato2018spectral}, hinge loss~\cite{lim2017geometric}\cite{tran2017deep}, projection discriminator~\cite{miyato2018cgans} and auxiliary classifier~\cite{odena2016conditional} into one model.
Machine learns the relationships between audio and visual information from watching videos.
We create a dataset from SoundNet Dataset \cite{aytar2016soundnet} by using pretrained image classification and sound classification models to apply data cleaning.
After training, the audio-to-image generator can produce recognizable images, and the advanced techniques of conditional GAN achieve better Inception score~\cite{salimans2016improved}\cite{barratt2018note} than the naive conditional GAN.
In addition, we show that our model learns the relationship between sounds and images by inputting the same sound with different volume levels.

\section{Related works}
Seeing and hearing help human to sense the world.
Some cross-modal researches try to learn the relation between auditory contents and visual contents.
For example, it is possible to learn  the relation between image and recorded spoken language sound~\cite{harwath2017learning}.
By computing similarity score between acoustic and visual features, the model can show that the specific object is attended when corresponding word is being told.
Moreover, the learning of neural network embeddings for natural images and speech waveforms describing the content of those images is explored~\cite{harwath2018vision}.
With natural image embedding as an interlingua, their experiments show that proposed models are capable of performing semantic cross-lingual speech-to-speech retrieval.

Sounds can not only interact with visual contents, sounds itself contain lots of information.
SoundNet~\cite{aytar2016soundnet} is a deep convolutional neural network for natural sound recognition.
By transferring the knowledge from other pretrained scene recognition model and object recognition model, SoundNet  learns to identify scenes and objects by only auditory contents.
Besides doing sound classification, information in sounds can also improve performance of other tasks, such as video captioning\cite{chuang2017seeing}~\cite{hori2017attention}\cite{hori2018multimodal}.
By adding sound features into video captioning models, the models generate more accurate descriptions and obtain higher scores in various evaluation metrics.

Recently there are lots of researches related to generative adversarial networks (GANs) \cite{goodfellow2014generative}.
In text-to-image~\cite{reed2016generative}, they turn a description into a vector representation first, and use this representation as input to generator.
They defined different losses to three different kinds of input pairs respectively.
After minimizing those losses, generator is capable of generating different kinds of images according to input text description.
Moreover, if some words are replaced by other words, just as the case that color is replaced from 'red' to 'blue', generated image will also changed the color from red to blue.

Besides generating images from given conditions, there are researches generating sounds from given videos, such as \cite{owens2016visually}.
In this research, they use deep convolution network to extract features from video screenshots.
Then, input these features into LSTM\cite{hochreiter1997long} to generate waveform that is correspond to input video.

There are some similar works that generate images condition on sounds, such as\cite{chen2017deep}\cite{hao2017cmcgan}.
In these works, they use different dataset called Sub-URMP\cite{chen2017deep}\cite{li2016creating} which is composed of sounds of musical performances with monotonous background and similar composition in images.
By using different training scenario, they achieve the goal of generating images which depict a single person with an instrument correspond to input sound.
However, in our work, we want to know whether machines can generate more complicated images condition on more complicated sounds.
Some different experimental results will be shown in section 5.

\begin{figure*}[t]
\centering
\includegraphics[width=0.85\textwidth]{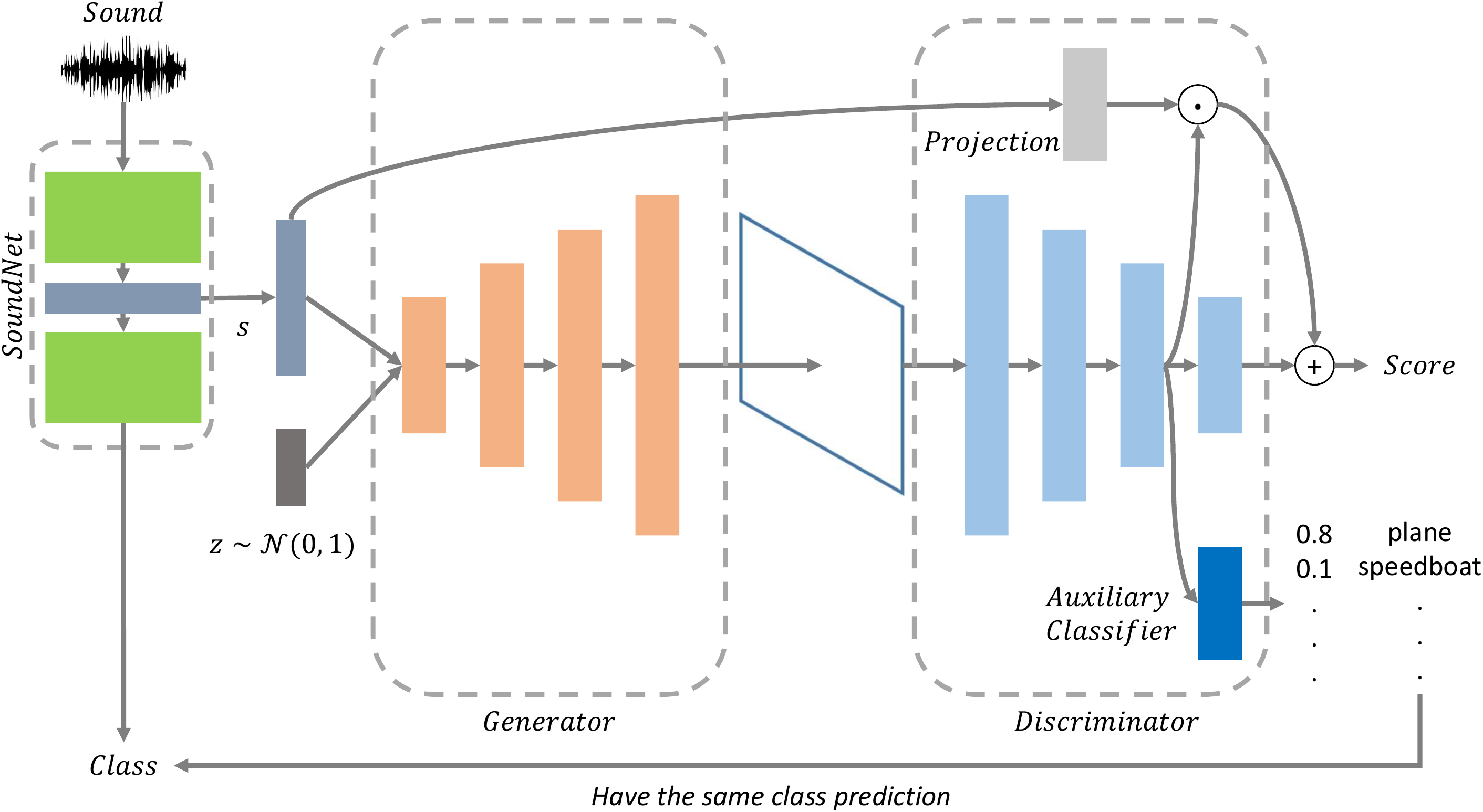}
\caption{Model architecture with projection discriminator and auxiliary classifier.}
\label{fig:model_architecture}
\end{figure*}

\section{Dataset} 
\label{sec:dataset}


To make machines learn the relation between sounds and images, we need paired sounds and images.
Fortunately, this kind of paired data can be collected easily.
In the videos that were recorded via cell phones or cameras, sound and image are highly correlated, and can be seen as paired data.
In previous work \cite{aytar2016soundnet}, videos crawled from the webs are used to train a sound classification model, SoundNet, to classify where or what is in the sounds.
Here we use the screenshots of videos and sound segment files in the dataset to train our audio-to-image models.
Most of sound segment files in our dataset are around 30 seconds long, and we resize all the screenshots to  size of 64*64.

However, we found that the corpus for training SoundNet cannot be directly used to train audio-to-image models because there are some discrepancies between images and sounds.
The screenshots and the sounds of videos can be unrelated.
For example, from the sound of video, we can hear sound of boat engine and rippling sound of water, but because the photographer was sit in the boat, we can only see the inside of the boat in the video. 
The discrepancies above may lead machines to learn chaotic relation between sounds and images.
In addition, the sounds of different objects sometimes cannot be discriminated even by humans.
For example, the sounds of boat engine are very similar to the sounds of propeller aircraft engine.
Because the model cannot discriminate their sounds,  when hearing the sounds of aircraft engine, the generator learned without data cleaning may generate the photo showing blue ocean with some splashes rather than the photo of plane flying.

To relieve the difficulties of learning sound-image matching, we use an image classifier and a sound classifier to clean up the dataset automatically.　
We classified sounds in those videos into categories by the pretrained sound classification model, SoundNet~\cite{aytar2016soundnet}.
We also use Inception model\cite{szegedy2016rethinking}, an image classification model, to classify the images.
If the classification results for the image and sound are not the same, the sound-image pair would be discarded.
After this procedure, 78\% of the data is discarded.
Because the above data cleaning procedure is automatic, it cannot be perfect, but it remarkably improves the quality of the generation results.
Because some objects are very rare in the training data, to make the training of audio-to-image plausible, only the sounds classified into dog, drum, guitar, piano, plane, speedboat, dam, soccer, baseball by SoundNet are used in the following experiments. 
The above nine classes are chosen because they are the classes with the most examples in the training data.
The number of training examples for each class is listed in Table~\ref{table:different_classes_training_data}.
The total number of sound-image pairs for  training is 10701, and the total number of sound segments for testing is 248.

\begin{table}[hb]
\centering
\begin{tabular}{|c|c||c|c|}
\hline
Class & \# of data & Class & \# of data\\
\hline\hline
Plane & 2803 & Speedboat & 900\\
\hline
Guitar & 207 & Piano & 1899\\
\hline
Drum & 259 & Dog & 264\\
\hline
Dam & 584 & Baseball & 1708\\
\hline
Soccer & 2077 & &\\
\hline
\end{tabular}
\caption{Number of training data in different classes.}
\label{table:different_classes_training_data}
\end{table}

\section{Approach}
\label{sec:approach}
Given pairs of sound segments and images, an audio-to-image generator is learned.
Due to the success of text-to-image synthesis~\cite{reed2016generative}, which utilized text embeddings as  condition for generators to generate correlated images, our work is based on similar model architecture. 
Recently, there are some researches trying to improve the generation by limiting discriminator to be a function in 1-Lipschitz continuity \cite{miyato2018cgans}\cite{miyato2018spectral}\cite{arjovsky2017wasserstein} or utilizing another auxiliary classifier in discriminator \cite{odena2016conditional}.
We fuse these approaches into one model. 
Therefore, although the algorithm for GAN training is similar to text-to-image\cite{reed2016generative}, the discriminator architecture and loss function used here are very different.
The model architecture is illustrated in Figure \ref{fig:model_architecture}.

\subsection{Generator}
The generator is shown in the left hand side of Figure \ref{fig:model_architecture}.
The input sound segment is first represented by a sequence of features.
The features can be spectrograms, fbanks, and mel-frequency cepstral coefficients (MFCCs), and the hidden layer outputs of the pretrained SoundNet model. 
Using SoundNet for feature extraction is illustrated in Figure \ref{fig:model_architecture}.
Then all the features in the sequence are averaged into a single vector　$s$.
The vector $s$ is taken as the condition of the generator.
Then, we concatenate a noise vector $z$ sampled from normal distribution with our sound condition as the input to generator. 
Generator is the cascade of several transposed convolution layers with hyperbolic tangent function as the activation function in the last layer. 
The output of the generator is an image generated based on the input condition.

\subsection{Discriminator}
The discriminator is in the right hand side of Figure \ref{fig:model_architecture}.
The discriminator takes a pair of sound segment and image as input, and outputs a score.
The architecture of discriminator is the cascade of several convolution layers with spectral normalization~\cite{miyato2018spectral} in each layer.
The convolution layers takes an image as input and outputs a scalar representing the quality of the image.
The projection layer which is simply a linear transformation  projects the sound vector into a latent representation~\cite{miyato2018cgans}. 
Then by computing inner-product between projected vector and the output of one of the convolution layer, we obtain a similarity score representing the degree of match between the audio and image.
The final output of the discriminator is the addition of the similarity score and the scalar that solely comes from convolution layers.
The final score represents  not only the realness of images but also relevance between sounds and images.
The discriminator learns to assign large score to the sound-image pairs in the training data, and low score to the sound and its generated image.
While the generator tries to fool discriminator, it learns how to generate images which are relevant to input condition and looks like real photos.

In Figure \ref{fig:model_architecture}, there is an auxiliary classifier.
The classifier shares weights with the convolution layers in discriminator, and they are jointly learned.
Because in the training data, the class of the sound segment and image pair can be obtained by SoundNet and Inception model, the classifier can learn to predict the class of an input image from the training data.
The generator will learn to generate images that can be correctly classified by the auxiliary classifier.
That is, given the sound segment that is classified as ``speedboat'' by SoundNet, the generator should generate the image that is also been classified as ``speedboat'' by the auxiliary classifier.

\subsection{Training Algorithm}
The loss functions of generator $G$ and discriminator $D$ are as follows.

Loss function of generator $\mathcal{L}_{G}$:
\begin{equation}
\begin{split}
\mathcal{L}_{G}= -\mathbb{E}_{ s \sim data, c = SN(s), z \sim \mathcal{N}(0,1) }
[ D(G(s,z), s) \\
+ \log P_C(c|G(s,z)) ].
\end{split}
\label{eqn:loss_of_generator}
\end{equation}
$s$ is the vector representation of a sound segment sampled from training data.
$SN(.)$ represents the SoundNet, and $c$ is the output class of the input sound $s$.
$G(s,z)$ is the generated image given sound $s$ and a random noise $z$ sampled from normal distribution.
$D(G(s), s)$ is the score assigned by the discriminator $D$ given a pair of sound $s$ and image $G(s)$.
The generator learns to maximize the score that can be obtained by the generated image $G(s)$.
$P_C(.)$ represents the auxiliary classifier.
The generator also learns to maximize the log likelihood of the auxiliary classifier, $\log P_C(c|G(s, z))$.

Loss function of discriminator $\mathcal{L}_{D}$:
\begin{equation}
\begin{split}
\mathcal{L}_{D}=
&   \mathbb{E}_{(s,x)\sim data}  [ \max(0, 1 - D(x, s)) ] \\
& + \mathbb{E}_{s\sim data,z \sim \mathcal{N}(0,1)} [ \max(0, 1 + D(G(s,z), s))] \\
& - \mathbb{E}_{x\sim data, c=IN(x)} [ \log P_C(c|x) ]
\end{split}
\label{eqn:loss_of_discriminator}
\end{equation}
In the first term, a pair of sound $s$ and image $x$ is sampled from the dataset.
The discriminator $D$ learns to assign larger score $D(x, s)$ to the pair to minimize $\mathcal{L}_{D}$.
Here we use hinge loss which is shown to improve the performance in the following experiments \cite{lim2017geometric}\cite{tran2017deep}.
In the second term, the sound $s$ is sampled from training data, while $G(s,z)$ is the image generated by the generator.
The discriminator leans to assign smaller score to the generated images.
In the third term, we sample a sound segment $s$ from the data set, and obtain its class $c$ by the Inception model $IN(.)$.
The auxiliary classifier $P_C$ learns to maximize the log likelihood $\log P_C(c|x)$ of class $c$ given image $x$.



The generator and the discriminator are trained iteratively.
That is, the generator is fixed, and the discriminator is updated several times to minimize $\mathcal{L}_D$.
Then we fix the discriminator, and update the parameters of the generator also several times to minimize $\mathcal{L}_G$.

\section{Experiments}
\label{sec:Experiments}



Our training procedure follows standard GAN training algorithm.
Generator is composed of four deconvolution layers with ascending number of kernels.
Discriminator is composed of four convolution layers and with linear function as activation function of final layer.
It is common to update the generator and discriminator with different numbers of steps.
According to our experimental observations, the number of training steps for discriminator in one iteration need to be less than generator.
It is easier to discriminate the authenticity of images and relevance between sounds and images than generate a real image.
To keep this adversarial training procedure in balance, more training steps are needed for generator to catch up discriminator.
We train generator five times per each update of discriminator.
The input dimension is 266 which consist of 256-dimension SoundNet feature and 10-dimension $z$ sampled from normal distribution.
The whole optimization process is based on Adam optimizer with learning rate 0.0002, and we train 300 epochs for all experiments.

\subsection{Sound Feature Representation}
First of all, we want to know which kind of sound feature is the most suitable feature for this task.
We use the Inception score to evaluate the generated images.
Inception score~\cite{salimans2016improved} is computed by extracting class distributions from generated images via pretrained image classification model Inception v3.
By feeding a generated image into Inception v3, we obtain a class distribution. 
If the class distribution is concentrated on one class, that means the image is clear, so Inception v3 is confidence about what it sees.
On the other hand, given a set of generated images, we want the average of the class distributions is more like uniform distribution because this means that the generated images are diverse.
Inception score integrates the above two properties into one score by using KL divergence.
Here the images  generated for testing are splitted into ten folds.
We calculated the Inception score for each fold, and show the  mean and standard deviation of the ten scores.


\begin{table}[htb]
\centering
\begin{tabular}{|c|c|}
\hline
Feature & Inception Score\\
\hline
Spectrogram & 2.16 $\pm$ 0.29\\
\hline
Fbank & 2.12 $\pm$ 0.32\\
\hline
MFCCs & 1.21 $\pm$ 0.09\\
\hline
SoundNet & 2.70 $\pm$ 0.73\\
\hline
\end{tabular}
\caption{Inception scores of different kinds of features.}
\label{table:inception_score_of_different_features}
\end{table}

The Inception scores of different sound features using the same model and training algorithm are shown in Table ~\ref{table:inception_score_of_different_features}.
For SoundNet feature, we used the output of the 18-th hidden layer.
The results show that SoundNet feature performs the best, so we utilize SoundNet feature in the rest experiments.
Among all the features, MFCCs performs the worst.
This is probably because MFCC is designed for speech recognition, and it discards some information not related to speech.
Spectrogram and fbank outperforms MFCC because they are more primitive than MFCC, and  preserves more information in the input audio.

\subsection{Qualitative Results}
\begin{figure*}[ht]
\centering
\begin{tabular}{|c|c|c|}
\hline
Dog & Drum & Guitar\\
\hline
\includegraphics[width=0.078\linewidth]{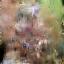}\includegraphics[width=0.078\linewidth]{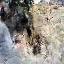}\includegraphics[width=0.078\linewidth]{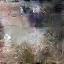}\includegraphics[width=0.078\linewidth]{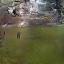} &
\includegraphics[width=0.078\linewidth]{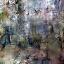}\includegraphics[width=0.078\linewidth]{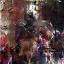}\includegraphics[width=0.078\linewidth]{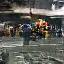}\includegraphics[width=0.078\linewidth]{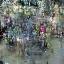} & \includegraphics[width=0.078\linewidth]{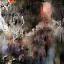}\includegraphics[width=0.078\linewidth]{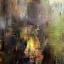}\includegraphics[width=0.078\linewidth]{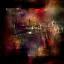}\includegraphics[width=0.078\linewidth]{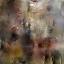} \\
\hline\hline
Piano & Plane & Speedboat\\
\hline
\includegraphics[width=0.078\linewidth]{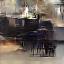}\includegraphics[width=0.078\linewidth]{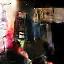}\includegraphics[width=0.078\linewidth]{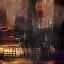}\includegraphics[width=0.078\linewidth]{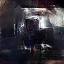} &
\includegraphics[width=0.078\linewidth]{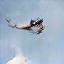}\includegraphics[width=0.078\linewidth]{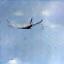}\includegraphics[width=0.078\linewidth]{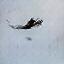}\includegraphics[width=0.078\linewidth]{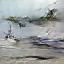} &
\includegraphics[width=0.078\linewidth]{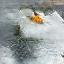}\includegraphics[width=0.078\linewidth]{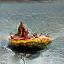}\includegraphics[width=0.078\linewidth]{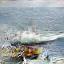}\includegraphics[width=0.078\linewidth]{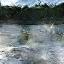}\\
\hline\hline
Dam & Soccer & Baseball\\
\hline
\includegraphics[width=0.078\linewidth]{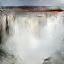}\includegraphics[width=0.078\linewidth]{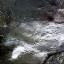}\includegraphics[width=0.078\linewidth]{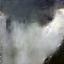}\includegraphics[width=0.078\linewidth]{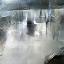} &
\includegraphics[width=0.078\linewidth]{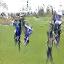}\includegraphics[width=0.078\linewidth]{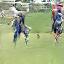}\includegraphics[width=0.078\linewidth]{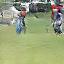}\includegraphics[width=0.078\linewidth]{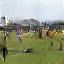} & \includegraphics[width=0.078\linewidth]{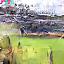}\includegraphics[width=0.078\linewidth]{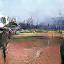}\includegraphics[width=0.078\linewidth]{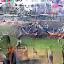}\includegraphics[width=0.078\linewidth]{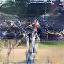}\\
\hline
\end{tabular}
\caption{Samples from our model. Each image is generated from a sound segment. The labels are the classes predicted by SoundNet.}
\label{fig:qualitative_samples_nine}
\end{figure*}

Sampled images from generator by inputting the sounds not in training data are shown in Fig \ref{fig:qualitative_samples_nine}. The audio files and their generated images can be found in \url{https://wjohn1483.github.io/audio_to_scene/index.html}.
The labels on top of the images are the class of input testing sounds predicted by SoundNet.
Although generated images are not as clear as normal photos, we can still see the shapes of some objects related to the input audio in some images.

Sounds belonging to some classes can generate relatively high quality images.
For speedboat or plane, there are eye-catching objects in the generated images. 
The generator truly generates the images that are interpretable to some extent.
Some classes of images get worse quality of images than others.
This may be because the imbalance and variance in different classes of training data.
The numbers of images in different classes are shown in Table \ref{table:different_classes_training_data}.
The number of training examples may explains why some classes performed better than the others.
We also found that for all the sounds classified into drums, they still have very high diversity.
There are many kinds of drum and are played in variant places.
It becomes an obstacle for model to generate image from the sound of such class.
On the contrary, in some classes like plane and speedboat with relatively common background such as blue sky and blue ocean, it is easier for model to generate high quality image in these classes.
In our dataset, we can assume that classes with natural landscape in background such as plane, speedboat, baseball, soccer, and dam are purer than classes with variant background such as dog, drum, guitar, and piano.

\subsection{Sound Volume}
To further investigate whether our model truly learns the relation between sound an vision, we tune the volume of sounds to observe the influences on generated images.
For example, if the sound is louder, the object may be closer or bigger in the generated image.
After tuning the volume of testing sounds, we extract previous mentioned SoundNet features for those sound files.
We input those tuned sound features into our generator which was pretrained on standard volume scale.

The images are shown in Table \ref{table:different_volume}.
The images in Table \ref{table:different_volume} are sampled from class speedboat and dam.
The images in the same row are generated from the same audio with different volumes.
The audio files can also be found in \url{https://wjohn1483.github.io/audio_to_scene/index.html}.
The numbers on top indicates the scale of volume that we modified our sound files.
In those images, we can see different scale of splashes.
As the volume goes up, the scale of splashes become larger.
We can see that our model truly learned the relation between characteristic of sound and image.
In this case, the volume of sounds is reflect on splashes.

\begin{table}[ht]
\centering
\begin{tabular}{|c|c|c|c|}
\hline
0.5 times & Original & 2 times & 3 times\\
\hline\hline
\includegraphics[width=0.2\linewidth]{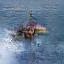} & \includegraphics[width=0.2\linewidth]{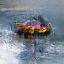} & \includegraphics[width=0.2\linewidth]{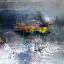} & \includegraphics[width=0.2\linewidth]{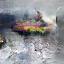}\\
\hline
\includegraphics[width=0.2\linewidth]{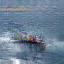} & \includegraphics[width=0.2\linewidth]{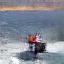} & \includegraphics[width=0.2\linewidth]{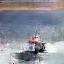} & \includegraphics[width=0.2\linewidth]{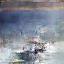}\\
\hline
\includegraphics[width=0.2\linewidth]{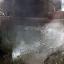} & \includegraphics[width=0.2\linewidth]{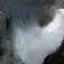} & \includegraphics[width=0.2\linewidth]{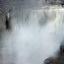} & \includegraphics[width=0.2\linewidth]{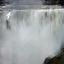}\\
\hline
\end{tabular}
\caption{Generated images by inputting different volumes of sounds. The numbers in the table is the relative loudness to the original sound.}
\label{table:different_volume}
\end{table}

\subsection{Ablation Study} \label{sec:ablation_study}
The architecture of our model contains spectral normalization, hinge version loss, projection discriminator and auxiliary classifier.
In this subsection, we want to know the influence of each part in our model.
Table \ref{table:ablation_study} shows the Inception score of different types of model.

\begin{table}[ht]
\centering
\begin{tabular}{|l|c|}
\hline
Model & Inception Score\\
\hline
(a) Upper bound & 4.44 $\pm$ 1.91\\
\hline
(b) Improved WGAN & 1.42 $\pm$ 0.13\\
\hline
(c) Conditional GAN & 2.21 $\pm$ 0.38\\
\hline
(d) +\ Spectral Norm & 2.45 $\pm$ 0.48\\
\hline
(e) +\ Hinge Loss & 2.49 $\pm$ 0.51\\
\hline
(f) +\ Projection Discriminator & 2.61 $\pm$ 0.41\\
\hline
(g) +\ Auxiliary Classifier & 2.83 $\pm$ 0.53\\
\hline
\end{tabular}
\caption{Inception scores of different models}
\label{table:ablation_study}
\end{table}

Row (a) shows the upper bound of this task, which is obtained by inputting all the real images we have in training and testing data to calculate Inception score.
The Inception score obtained in this way is 4.44, which is the highest score we can get. 
We can use this upper bound score as a criterion to measure the quality between generated images and real images. 

In both rows (b) and (c), we used the same network architecture as in \cite{reed2016generative}, but we substitute sound embedding for sentence embedding.
In row (b), we apply improved W-GAN\cite{arjovsky2017wasserstein} on original text to image architecture, which use gradient penalty to make sure discriminator is in 1-Lipschitz continuity.

The table shows that improved W-GAN cannot get good Inception score in this task.
On the other hand, conditional GAN can perform better.
By adding different tricks mentioned above, we can get improvements step by step.
It shows that tricks do help our model to generate better images.
Finally, with all the technologies, we can get 2.83 in Inception score, which performs relatively good compare to our upper bound.

\subsection{Human Evaluation}
\subsubsection{Evaluation on ablation study}
In the previous section, we use Inception score to evaluate the realness of generated images.
In this section, we want to prove that the improvement of different models is not only shown on Inception score but also on human feeling.
We ask ten people to help us evaluate our models.
Our experimental setup is as follows, we sample some pairs of image and corresponding sounds in testing data.
Then, let people listen to those testing sounds and rate from 1 to 5.
If the generated image is unreal or uncorrelated to testing sound, people should rate this pair with lower score.
On the contrary, if the generated image seems real enough and have high correlation with sound, this pair should get higher score.

\begin{table}[ht]
\centering
\begin{tabular}{|c|c|}
\hline
Model & Average Score\\
\hline
Conditional GAN with spectral norm & 1.90\\
\hline
+\ Hinge Loss & 2.74\\
\hline
+\ Projection Discriminator & 3.16\\
\hline
+\ Auxiliary Classifier & 3.70\\
\hline
\end{tabular}
\caption{Human scores on different models}
\label{table:human_ablation_study}
\end{table}

The results are shown in Table \ref{table:human_ablation_study}.
We can see that most people think the model with all tricks performed the best.
Although those models get close scores in Inception score, they get scores which have at least 0.4 gap between different models.

\subsubsection{Correlation between sounds and images}
In section \ref{sec:ablation_study}, we shows that our model can generate realistic images and relative good Inception score.
However, it only shows the realness of generated images.
To measure the correlation between sounds and images, we ask people to choose the most correlated image from two different images after hearing a sound from testing data.
These two images are conditioned on different class of sounds so that if our model can generate images related to given class, people will choose the corresponding image which is generated by inputting sound that they just listen to, rather than image generated by inputting sampled sound from other classes.

\begin{table}[ht]
\centering
\begin{tabular}{|c|c|c|c|}
\hline
Options & Positive & Negative & Neither\\
\hline
Percentage (\%) & 73 & 11 & 16 \\
\hline
\end{tabular}
\caption{Human scores on correlation between sounds and images}
\label{table:human_evaluation_correlation}
\end{table}

The results are listed in Table \ref{table:human_evaluation_correlation}.
Options in table means the choices that people choose.
Positive means people choose the image generated by the sound they hear, negative means people choose the image generated by sound sampled from other classes, and neither means people think both of the images cannot represent the sound they listen to.
We can see that most of the people think the images that our model generated are correlated to input sounds.
It shows that our model has the ability to generate images related to given sounds.
\vspace{-1mm}
\section{Conclusion}
\label{sec:conclusion}
In this paper, we introduce a novel task in which images are generated  conditioned on sounds.
Base on SoundNet dataset, we utilize image and sound classification results to build a relatively cleaner image-sound paired dataset.
By applying different methods to our generative model, the model can generate images with better quality in terms of both subjective and objective evaluations.
In addition, almost three-fourth of people agree that our model have the ability to generate images related to sounds.

In future work, we plan to use more wide variety of sounds to train our model.
Also, we plan to learn a model to generate sounds from visual information~\cite{owens2016visually}, and improve our model by dual learning strategy.
By applying dual learning, we can expect the performance of both models can gain improvements from matching loss and cycle consistency loss.


\bibliographystyle{IEEEbib}
\bibliography{refs}

\end{document}